\documentclass[a4paper, 10 pt, conference]{ieeeconf}

\IEEEoverridecommandlockouts                              

\usepackage{xcolor}

\usepackage{graphics} 
\usepackage{epsfig} 
\usepackage{mathptmx} 
\usepackage{amsmath} 
\usepackage{amssymb}  
\usepackage{textgreek}
\usepackage{hyperref}
\usepackage{stfloats}
\usepackage{flushend}
\usepackage{subfigure}
\usepackage{xspace}

\usepackage[belowskip=-10pt,aboveskip=12pt]{caption}
 
\title{\LARGE \bf 3D Vision-guided Pick-and-Place Using Kuka LBR iiwa Robot}

\author{Hanlin Niu, Ze Ji, Zihang Zhu, Hujun Yin, and Joaquin Carrasco
\thanks{*This work was supported by ASTUTE 2020 (Advanced Sustainable Manufacturing Technologies), EPSRC project No.EP/S03286X/1 and Royal Society (RGS/R1/201159 research grant). \textit{(Corresponding author: Hanlin Niu)}}
\thanks{H. Niu, H. Yin and J. Carrasco are with the Department of Electrical \& Electronic Engineering, The University of Manchester, Manchester, UK.
        {\{\tt\small hanlin.niu@manchester.ac.uk}\}}
\thanks{Z. Ji is with the School of Engineering, Cardiff University, Cardiff, UK. }
\thanks{Z. Zhu is with Körber Digital GmbH, Körber AG, Germany.}

}

\begin{document}

\maketitle

\begin{abstract}

This paper presents the development of a control system for vision-guided pick-and-place tasks using a robot arm equipped with a 3D camera. The main steps include camera intrinsic and extrinsic calibration, hand-eye calibration, initial object pose registration, objects pose alignment algorithm, and pick-and-place execution. The proposed system allows the robot be able to to pick and place object with limited times of registering a new object and the developed software can be applied for new object scenario quickly. The integrated system was tested using the hardware combination of kuka iiwa, Robotiq grippers (two finger gripper and three finger gripper) and 3D cameras (Intel realsense D415 camera, Intel realsense D435 camera,  Microsoft Kinect V2). The whole system can also be modified for the combination of other robotic arm, gripper and 3D camera.
\end{abstract}

\section{Introduction}

In recent years, there have been increasing interests with regard to flexible manufacturing systems that can adapt automation equipment to varying environments effectively in order to improve the productivity of a factory automation system. The implementation of a robotic arm can reduce the lead time and cost of manufacturing. Robotic arms can be used for multiple industrial applications, such as assembly, welding, material handling, painting and drilling. To establish a flexible manufacturing system in a production site, intelligent robots which can perceive the environment \cite{song20193d}, recognize the situation, and manipulate objects autonomously, are necessary. Pick-and-place systems that perform pick-and-place tasks for randomly placed and oriented parts from bins or boxes are considered a necessary part for such a manufacturing process. Despite considerable advancement in related areas, autonomous pick-and-place remains a challenge because it merges several specific robotic issues such as camera intrinsic and extrinsic calibration, hand-eye calibration, object recognition and localization, and grasp planning, into one common goal. To satisfy the industrial standard, the holistic system should be robust to the cluttered, random, dynamic, and dimly lit environment \cite{wu2016architecture} \cite{belzile2019workspace}.

\begin{figure}[h]
  \begin{center}
    \includegraphics[width=3.3in, height=2.0in]{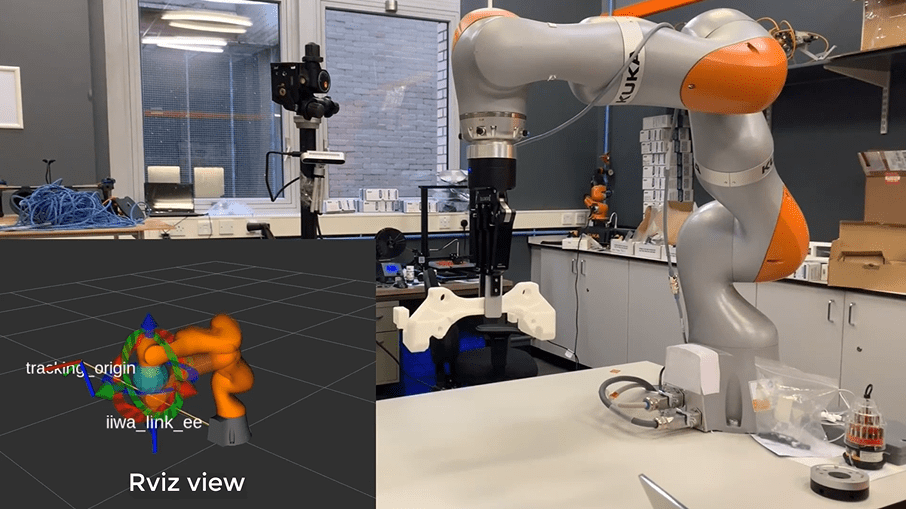}
    \caption{Randomly pick and place 3D printed object using Kuka iiwa robotic arm, Robotiq 2 finger gripper and Intel Realsense camera.}
    \label{fig1}
  \end{center}
\end{figure}

The goal of this research is to develop the whole software solution for the robot arm to pick and plac objects using a simple object registration process instead of using dedicated CAD model of the object, as shown in Fig.~\ref{fig1}. 

\section{Hardware and Software System}
In this research, a Kuka iiwa robot is implemented in the proposed system due to its large Robot Operating System (ROS) supporting community. Both Robotiq 2 finger gripper and 3 finger gripper are tested. Point cloud of objects are obtained using 3D cameras, including Intel realsense D415, D435, and Microsoft Kinect V2 cameras. 

\subsection{Design and manufacturing of universal adapter}
A universal adapter that can connect both grippers to the end effector of robot arm is designed and manufactured instead of using the default adapter for each gripper, which saves time to disassemble and assemble gripper.

\begin{figure}[ht]
\centering
\subfigure[]{\label{fig21}
\includegraphics[width=0.23\textwidth,height=0.18\textwidth]{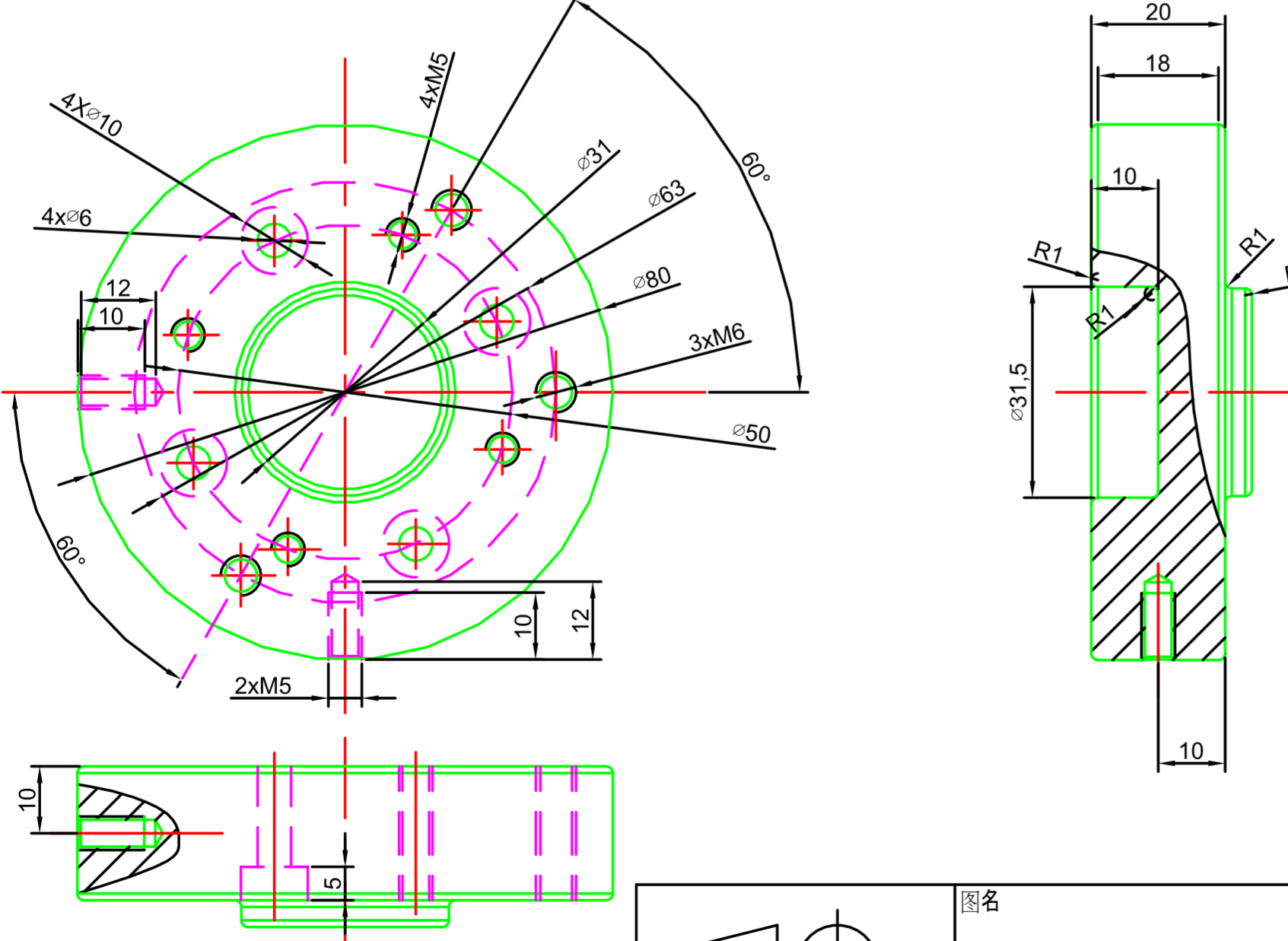}}
\subfigure[]{\label{fig22}
\includegraphics[width=0.23\textwidth,height=0.18\textwidth]{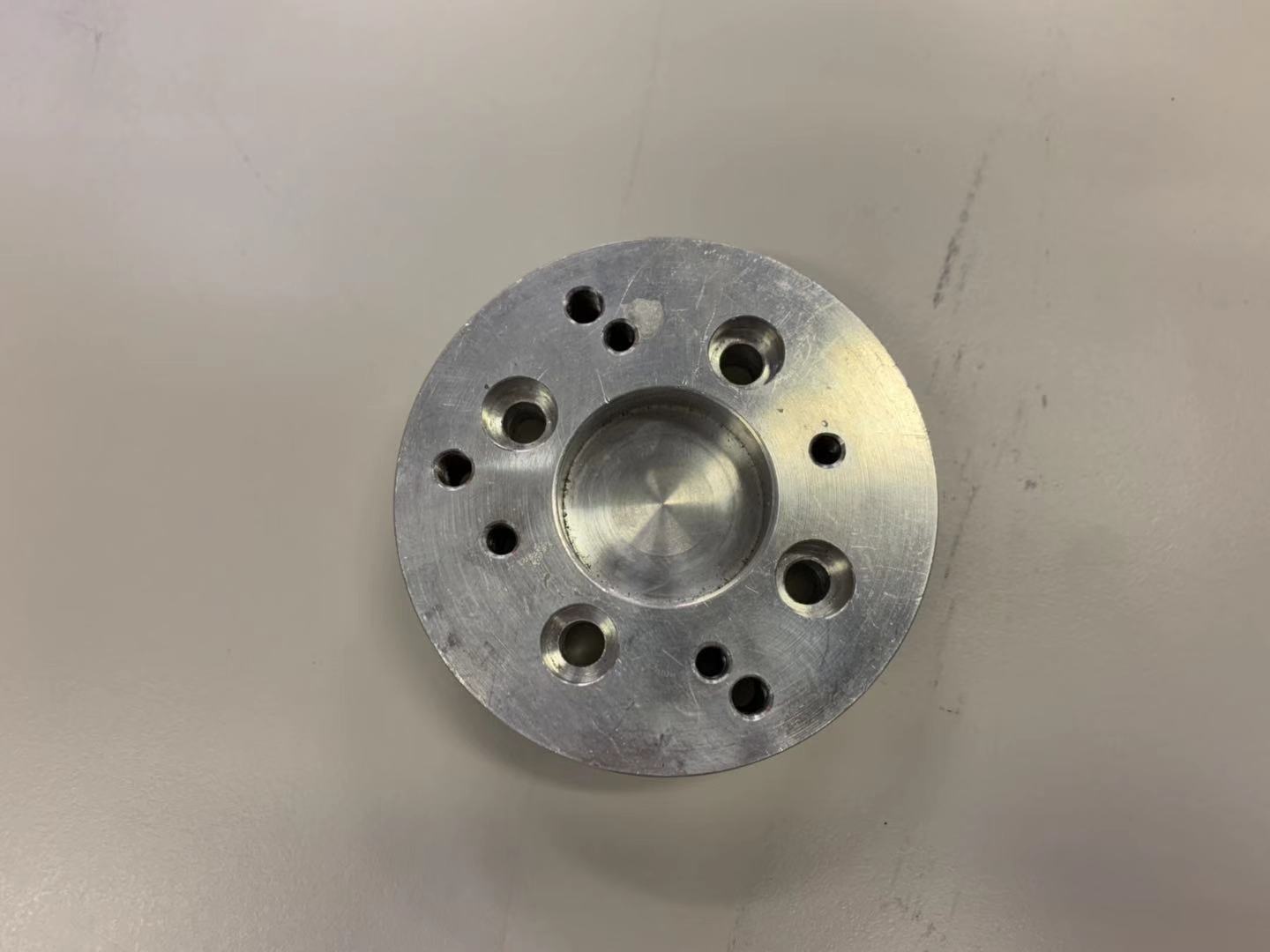}}
\caption{Design and manufacturing of gripper adapter: (a) CAD design (b) manufactured aluminium adapter.}
\label{fig2}
\end{figure}

\subsection{Camera intrinsic and extrinsic calibration}
\begin{figure}[h]
\centering
\subfigure[]{\label{fig33}
\includegraphics[width=0.23\textwidth,height=0.18\textwidth]{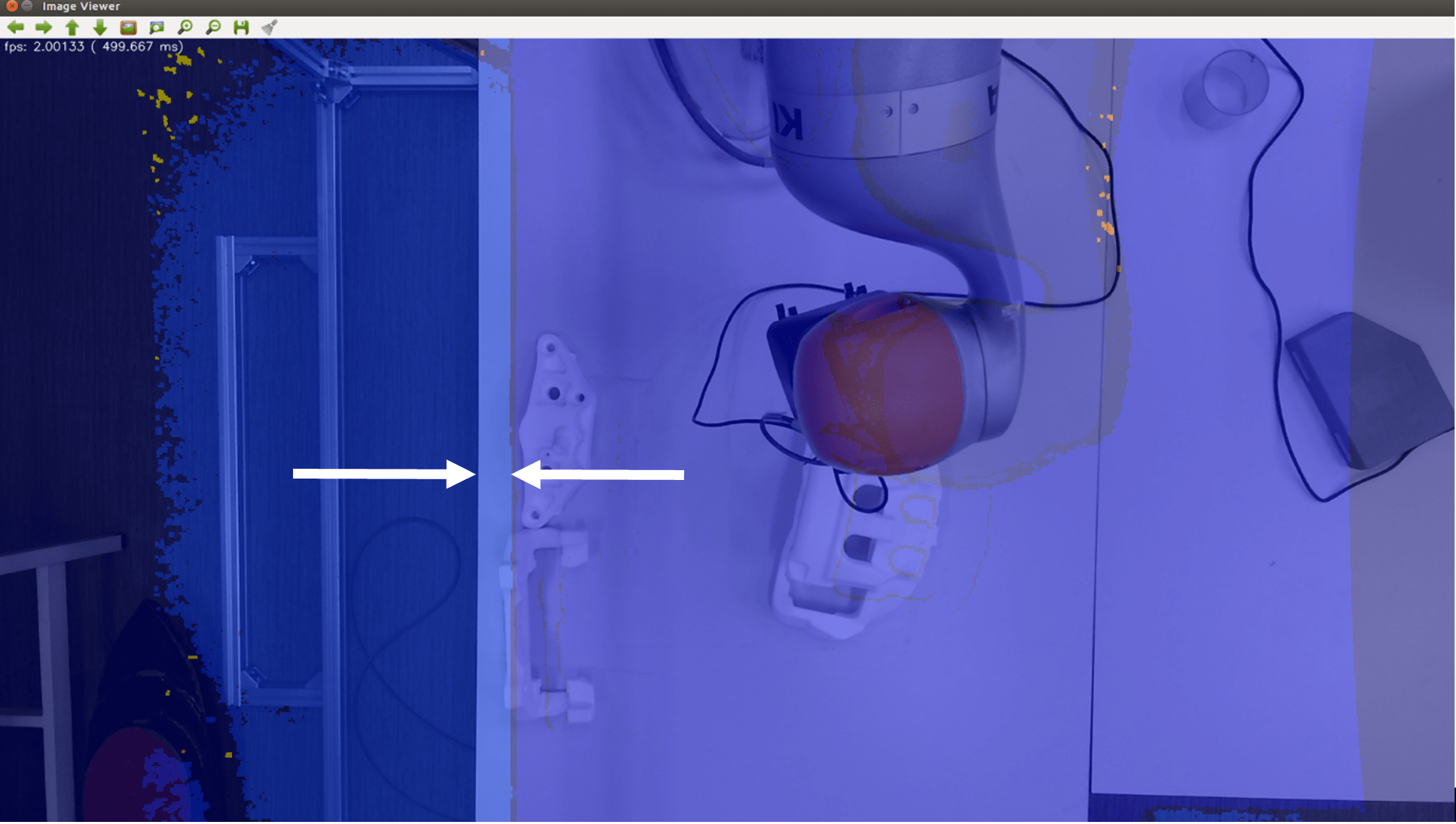}}
\subfigure[]{\label{fig34}
\includegraphics[width=0.23\textwidth,height=0.18\textwidth]{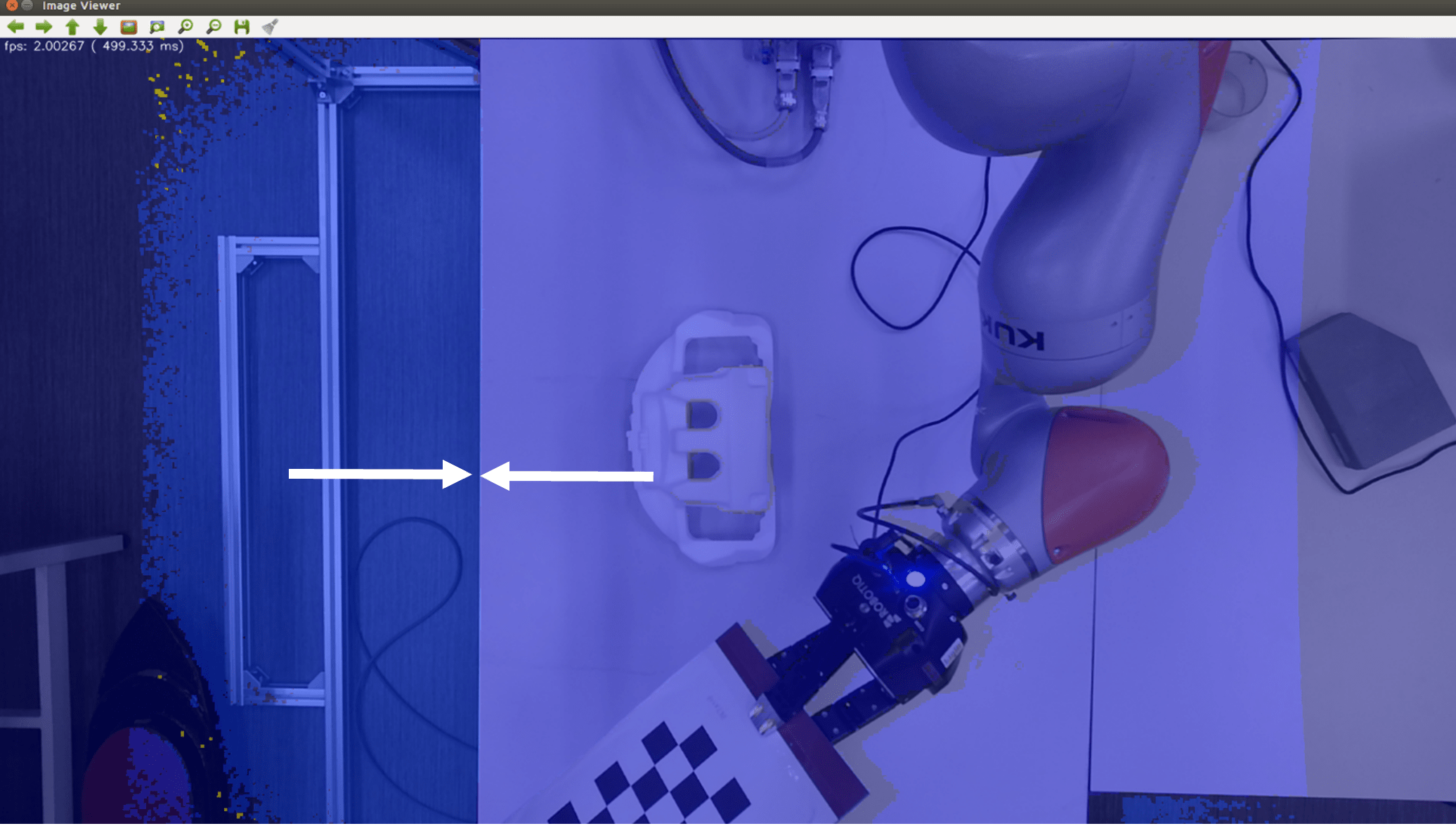}}
\caption{Camera accuracy (a) before and (b) after calibration: the gap between two while arrows represents the derivation of the camera depth data.}
\label{fig3}
\end{figure}

To achieve accurate scanning, intrinsic and extrinsic parameters of the Microsoft Kinect V2 camera need to be calibrated before using. The calibration software package can be found in \cite{iai_kinect2}. The intrinsic of the color camera and IR camera was calibrated by using 100 color images and 100 IR images of a chess board, respectively. The extrinsic of the camera was calibrated by using 100 synchronized images of both cameras. The derivation of the camera depth data and real data before calibration is illustrated in Fig.~\ref{fig33}. The gap between two white arrows represent the derivation. We can see that the gap diminish after calibration using 300 images of chess board. Note that Intel realsense 3D cameras are factory calibrated, it is unnecessary to recalibrate.


\subsection{Hand-eye calibration}

\begin{figure}[t]
\centering
\subfigure[]{\label{fig41}
\includegraphics[width=0.23\textwidth,height=0.18\textwidth]{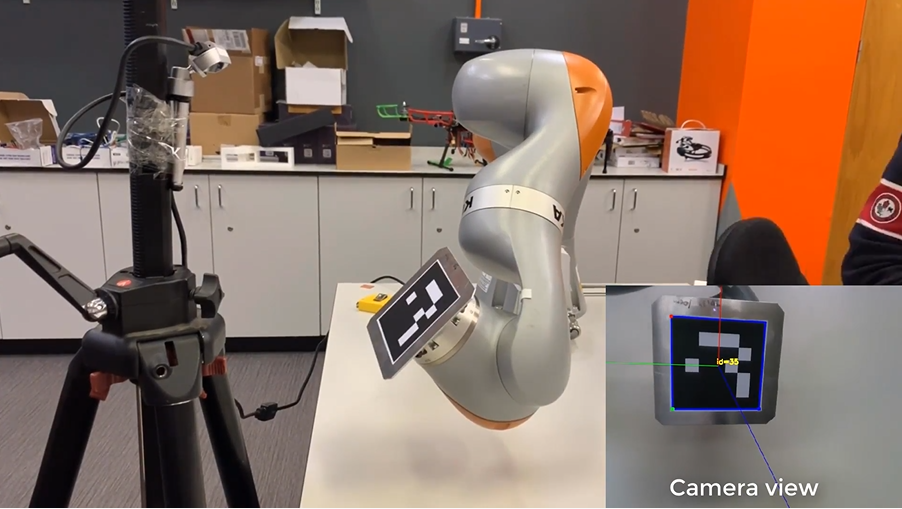}}
\subfigure[]{\label{fig42}
\includegraphics[width=0.23\textwidth,height=0.18\textwidth]{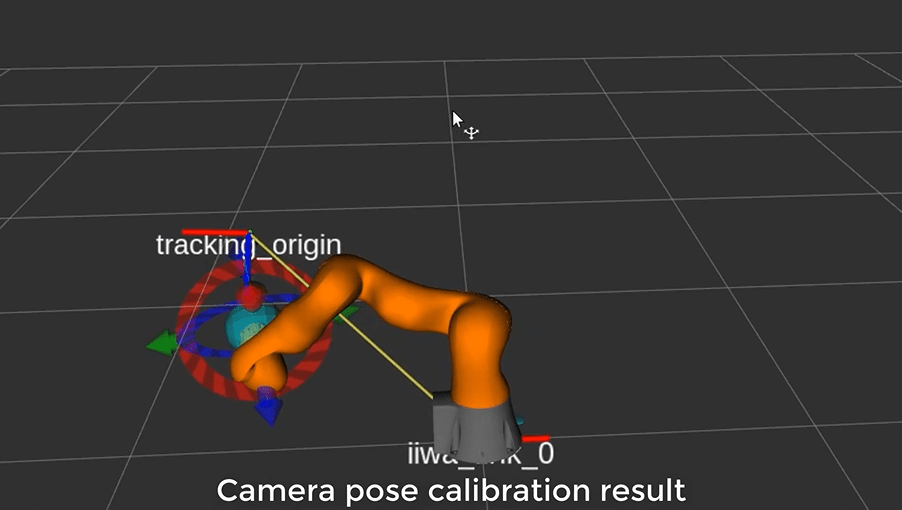}}
\caption{Hand-eye calibration: (a) calibration using AR marker (b) calibration result shown in Rviz.}
\label{fig4}
\end{figure}

The transformation matrix between the camera and the end-effector of the robot arm is achieved by detecting an Aruco (AR) marker, which is attached to the front end of robot arm, as shown in Fig.~\ref{fig41}. Fifteen sets of relative pose data and robot arm pose data were recorded to calculate the transformation matrix between the robot arm base and the camera. The calibration result is shown in Fig.~\ref{fig42}. The camera pose is represented by $tracking\_origin$ and robot arm base pose is illustrated by $iiwa\_link\_0$.

\subsection{Pose registration and pose alignment algorithm}

\begin{figure}[h]
\centering
\subfigure[]{\label{fig51}
\includegraphics[width=0.23\textwidth,height=0.18\textwidth]{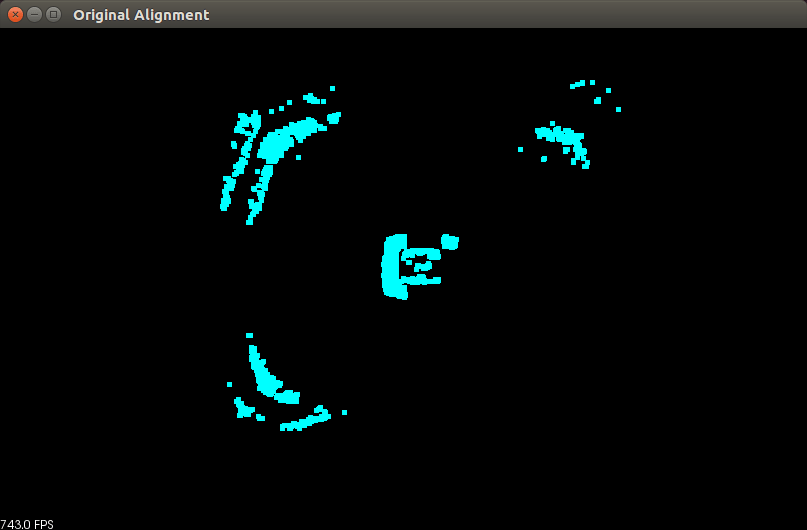}}
\subfigure[]{\label{fig52}
\includegraphics[width=0.23\textwidth,height=0.18\textwidth]{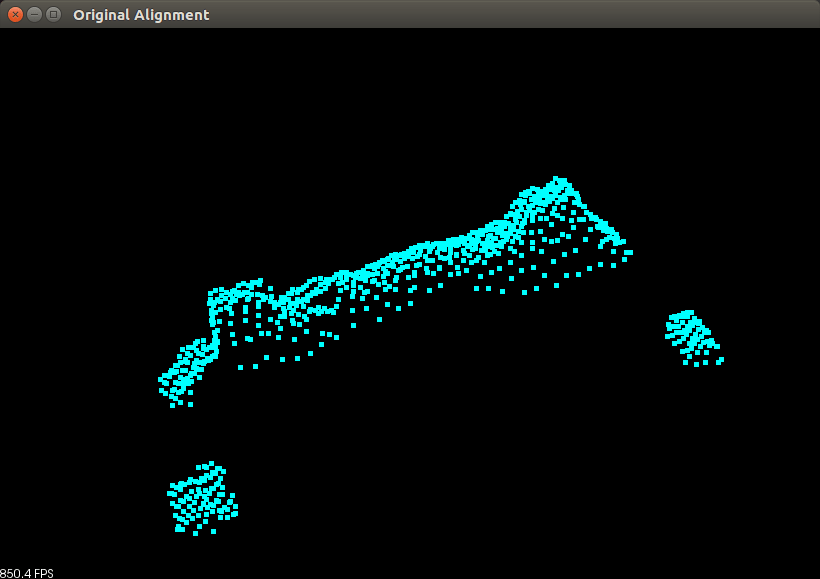}}
\subfigure[]{\label{fig53}
\includegraphics[width=0.23\textwidth,height=0.18\textwidth]{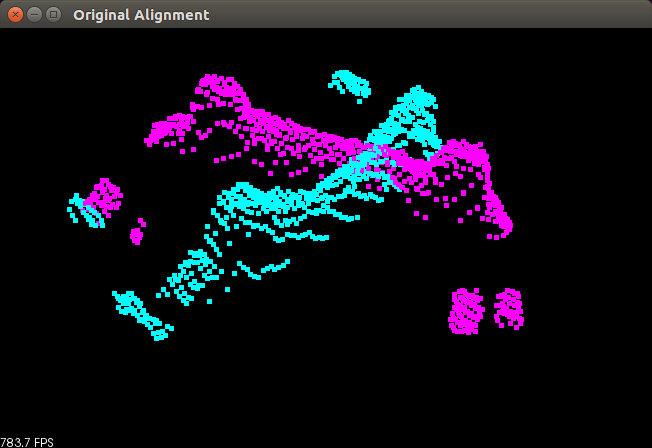}}
\subfigure[]{\label{fig54}
\includegraphics[width=0.23\textwidth,height=0.18\textwidth]{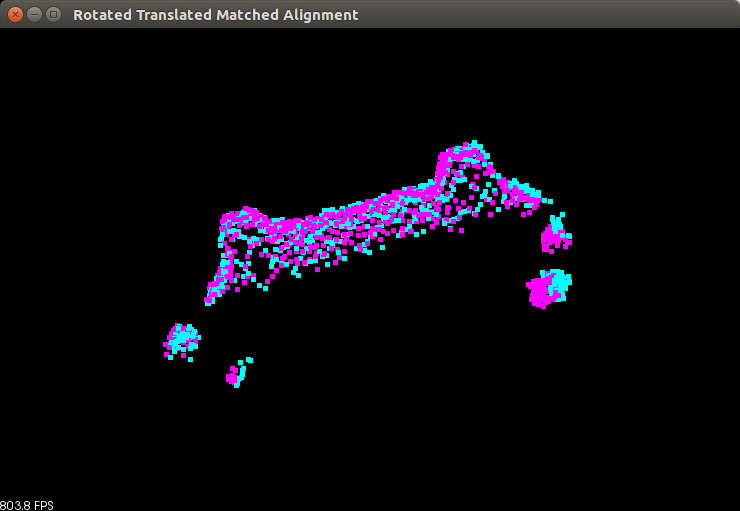}}
\caption{Pose registration and pose alignment algorithm: (a) point cloud before cropping and (b) after cropping (c) point cloud of registered pose and scanned pose (d) matching result}
\label{fig5}
\end{figure}

Each object is scanned for registration. The object point cloud is segmented from the background, as shown in Fig.~\ref{fig51} and Fig.~\ref{fig52}. For each registered segmented point cloud, the robot arm will be operated to a corresponding grasping pose. The pair of registered object point cloud and grasping pose will be recorded. After each face of the object is registered, the object can be placed randomly on the table. The random pose and the registered pose are shown in Fig.~\ref{fig53}. A pose alignment algorithm is used to calculate the transformation matrix between the registered pose and the random pose. For accuracy validation, the transformation matrix is used to transform the registered pose to the random pose. As shown in fig.~\ref{fig54}, the transformed registered pose can match with the random pose precisely. As each registered pose corresponds to a grasping pose, the grasping pose of the randomly placed object can also be calculated.

\subsection{Pick-and-place execution}
After calculating the pose of randomly placed an object, the kuka iiwa robot arm is commanded to pick the object from a randomly placed position and then place the object at a pre-defined unloading position. The combination of Robotiq two finger gripper and realsense cameras (D415 and D435) and the combination of Robotiq three finger gripper and Microsoft Kinect v2 camera are both compatible with the developed system, as shown in demonstration video.

\bibliographystyle{IEEEtran}
\bibliography{IEEEexample}
\end{document}